\documentclass[11pt,a4paper]{article}
\usepackage[hyperref]{acl2020}

\usepackage{amsmath}
\usepackage{amssymb}
\usepackage{multirow}
\usepackage{booktabs}
\usepackage{appendix}
\usepackage{enumitem}
\usepackage{graphicx}

\usepackage{mysty}
\usepackage{times}
\usepackage{latexsym}

\usepackage{microtype}

\aclfinalcopy % Uncomment this line for the final submission
\newif\ifshowcomment
\showcommenttrue

\title{A Better Variant of Self-Critical Sequence Training}

\author{Ruotian Luo \\
  TTI-Chicago \\
  \texttt{rluo@ttic.edu}}

\date{}

\begin{document}
\maketitle
\begin{abstract}
In this work, we present a simple yet better variant of Self-Critical Sequence Training. We make a simple change in the choice of baseline function in REINFORCE algorithm. The new baseline can bring better performance with no extra cost, compared to the greedy decoding baseline.
\end{abstract}

\section{Introduction}
Self-Critical Sequence Training(SCST), upon its release, has been a popular way to train sequence generation models. While originally proposed for image captioning task, SCST not only has become the new standard for training captioning models~\cite{yao2017boosting,dognin2019adversarial,luo2018discriminability,jiang2018recurrent,liu2018show,ling2017teaching,liu2018context,chen2019improving,yang2019auto,zhao2018multi}, but also has been applied to many other tasks, like video captioning\cite{li2018jointly,chen2018less,li2019end}, reading comprehension\cite{hu2017reinforced}, summarization\cite{celikyilmaz2018deep,paulus2017deep,wang2018reinforced,pasunuru2018multi}, image paragraph generation\cite{melas2018training}, speech recognition\cite{zhou2018improving}.

SCST is used to optimize generated sequences over a non-differentiable objective, usually the evaluation metrics, for example, CIDEr for captioning, ROUGE for summarization. To optimize such objective, SCST adopts REINFORCE with baseline~\cite{williams1992simple}, where a ``Self-Critical'' baseline is used; specifically, the score of the greedy decoding output is used as the baseline. This is proved to be better than learned baseline function which is more commonly used in Reinforcement Learning literature.

In this work, we present a different baseline choice which was first proposed in \cite{mnih2016variational}, to the best of our knowledge. With more elaboration in Sec. \ref{sec:nsc}, this baseline can be described as a variant of ``Self-Critical''. This method is simple, but also faster and more effective compared to the greedy decoding baseline used in SCST.

\section{Recap for SCST}

MIXER~\cite{ranzato2015sequence} is the first to use REINFORCE algorithm for sequence generation training.
They use a learned function approximator to get the baseline.

SCST inherits the REINFORCE algorithm from MIXER, but discards the learned baseline function. Instead, SCST uses the reward of the greedy decoding result as the baseline, achieving better captioning performance and lower gradient variance. Many variants of SCST have also been proposed~\cite{anderson2018bottom,gao2019self,hu2017reinforced}.

% While greedy decoding usually provide a good baseline, and proved to reduce the gradient variance compared to learned baseline functions, this can be improved. A common practice is to use value function(\rtcom{?}) as baseline, which is defined as expected reward. Our variant tries to estimate expected reward and use that as baseline.

\subsection{Math formulation}
The goal of SCST, for example in captioning, is to maximize the expected CIDEr score of generated captions.
$$\max_{\theta} E_{{\hat c}\sim p_{\theta}(c|I)}[R({\hat c})] $$
where ${\hat c}$ is a sampled caption; $I$ is the image; $p_{\theta}(c|I)$ is the captioning model parameterized by $\theta$, and $R(\cdot)$ is the CIDEr score.

Since this objective is not non-differentiable with respect to $\theta$, back propagation is not feasible. To optimize it, a policy gradient method, specifically REINFORCE with baseline~\cite{williams1992simple} is used.
$$\nabla_{\theta} E[R] \approx (R(\hat c)-b) \nabla_{\theta} \log p_{\theta}(\hat c| I)$$

The policy gradient method allows estimating the gradient from individual samples (the right-hand side) and applying gradient ascent. To reduce the variance of the estimation, a baseline $b$ is needed, and $b$ has to be independent of $\hat c$.

In SCST, the baseline is set to be the CIDEr score of the greedy decoding caption, denoted as $c^*$. Thus, we have
$$\nabla_{\theta} E[R] \approx (R(\hat c) - R(c^*)) \nabla_{\theta} \log p_{\theta}(\hat c| I)$$

\section{The Better SCST}\label{sec:nsc}

The success of SCST comes from better gradient variance reduction introduced by the greedy decoding baseline. In our variant, we use the baseline proposed in \cite{mnih2016variational} to achieve even better variance reduction.

Following \cite{mnih2016variational}, we sample $K$ captions for each image when applying REINFORCE: ${\hat c}_1 \ldots {\hat c}_K$, ${\hat c}_k \sim p_{\theta}(c|I)$, 

The baseline for each sampled caption is defined as the average reward of the rest samples. That is, for caption $\hat c_k$, its baseline is 
\begin{align}
    b_k = \frac{1}{K-1}\sum_{j\neq k} R(\hat c_j)
\end{align}
Since each sample is independently drawn, $b_k$ is a valid baseline. The final gradient estimation is 
\begin{align}
    \nabla_{\theta} \approx \left (R(\hat c_k) - \frac{1}{K-1}\sum_{j\neq k} R(\hat c_j)\right ) \nabla_\theta\log p({\hat c}_k|I)
\end{align}
Note that, $b_k$ is an estimation of expected reward, which is similar to the learning objective of value functions in other Reinforcement Learning algorithms. The expected reward is usually a good baseline choice in that it can effectively reduce gradient variance. In Sec. \ref{sec:exp}, we show that our gradient variance is lower than SCST empirically.

It is still a ``Self-Critical'' baseline because the critic is still from itself: its other sampling results, instead of the greedy decoding result. Our variant can also been seen as a special case of \cite{gao2019self}. While SCST is {\it T}-step-maxpro, ours is {\it T}-step-sample, with the naming in~\cite{gao2019self}.

\section{Experiments}\label{sec:exp}
For all models, we first pretrain them using standard cross-entropy loss and then switch to Self-Critical training.

For a fair comparison, during Self-Critical stage, we always sample 5 captions for each image, same for both SCST and our variant.

All the experiments are done on COCO captioning dataset~\cite{lin2014microsoft}. The scores are obtained on Karparthy test split~\cite{karpathy2015deep} with beam search of beam size 5 if not explicitly noted.

\paragraph{Speed} Since no extra greedy decoding is needed, our method is slightly faster than SCST.
\paragraph{Performance on different model architectures}
We experiment with four different architectures. FC and Att2in are from SCST\cite{rennie2017self}. UpDown is from \cite{anderson2018bottom}. Transformer is adapted from \cite{vaswani2017attention} for captioning task.

Table \ref{tab:differentmodels} shows that our variant is better than SCST on all architectures, especially on Transformer.

\begin{table*}[htbp]\small
  \centering
    \begin{tabular}{lcccccccc}
          & \multicolumn{1}{c}{Bleu1} & \multicolumn{1}{c}{Bleu2} & \multicolumn{1}{c}{Bleu3} & \multicolumn{1}{c}{Bleu4} & \multicolumn{1}{c}{ROUGE\_L} & \multicolumn{1}{c}{METEOR} & \multicolumn{1}{c}{CIDEr} & \multicolumn{1}{c}{SPICE} \\
    \hline
    FC(SCST) & 74.7  & 57.8  & 43.0  & \textbf{31.7} & 54.0  & 25.2  & 104.5 & 18.4 \\
    FC(Ours) & \textbf{74.9} & \textbf{57.9} & \textbf{43.0} & 31.6  & \textbf{54.1} & \textbf{25.4} & \textbf{105.3} & \textbf{18.6} \\
    \hline
    Att2in(SCST) & 78.0  & 61.8  & 47.0  & 35.3  & 56.7  & 27.1  & 117.4 & 20.5 \\
    Att2in(Ours) & \textbf{78.4} & \textbf{62.1} & \textbf{47.3} & \textbf{35.6} & \textbf{56.9} & \textbf{27.3} & \textbf{119.5} & \textbf{20.7} \\
    \hline
    UpDown(SCST) & 79.4  & 63.3  & 48.6  & 36.7  & 57.6  & 27.9  & 122.7 & \textbf{21.5} \\
    UpDown(Ours) & \textbf{80.0} & \textbf{63.9} & \textbf{49.1} & \textbf{37.2} & \textbf{57.8} & \textbf{28.0} & \textbf{123.9} & \textbf{21.5} \\
    \hline
    Transformer(SCST) & 80.0  & 64.6  & 50.3  & 38.6  & 58.4  & 28.6  & 126.6 & 22.2 \\
    Transformer(Ours) & \textbf{80.7} & \textbf{65.6} & \textbf{51.3} & \textbf{39.4} & \textbf{58.7} & \textbf{28.9} & \textbf{129.6} & \textbf{22.8} \\
    \hline
    \end{tabular}%
  \caption{The performance of our method on different model architectures. The numbers are from authors' own implementation.}
  \label{tab:differentmodels}%
\end{table*}%

\paragraph{Different training hyperparameters}
Here we adopt a different training setting (`Long') for UpDown model. The `Long' setting (from \url{https://github.com/yangxuntu/SGAE}) uses a larger batch size and a longer training time. Table \ref{tab:differenttraining} shows that there is always a gap between our method and SCST which cannot be closed by longer training or a larger batch size.
\begin{table*}[htbp]\small
  \centering
    \begin{tabular}{lcccccccc}
          & \multicolumn{1}{c}{Bleu1} & \multicolumn{1}{c}{Bleu2} & \multicolumn{1}{c}{Bleu3} & \multicolumn{1}{c}{Bleu4} & \multicolumn{1}{c}{ROUGE\_L} & \multicolumn{1}{c}{METEOR} & \multicolumn{1}{c}{CIDEr} & \multicolumn{1}{c}{SPICE} \\
    \midrule
    UpDown+SCST & 79.4  & 63.3  & 48.6  & 36.7  & 57.6  & 27.9  & 122.7 & \textbf{21.5} \\
    UpDown+Ours & \textbf{80.0} & \textbf{63.9} & \textbf{49.1} & \textbf{37.2} & \textbf{57.8} & \textbf{28.0} & \textbf{123.9} & \textbf{21.5} \\
    \midrule
    UpDown(Long)+SCST & 80.3  & 64.5  & 49.9  & 38.0  & 58.3  & 28.4  & 127.2 & 21.9 \\
    UpDown(Long)+Ours & \textbf{80.4} & \textbf{64.7} & \textbf{50.1} & \textbf{38.1} & \textbf{58.4} & \textbf{28.5} & \textbf{127.9} & \textbf{22.0} \\
    \bottomrule
    \end{tabular}%
  \caption{The performance of UpDown model with SCST/Ours under two different hyperparameter settings.}
  \label{tab:differenttraining}%
\end{table*}%

\paragraph{Multiple runs}
Table \ref{tab:multipleruns} shows that our variant is consistently better than SCST with different random seeds. All the models use `Long' setting with UpDown model.

Specifically, we pretrain 5 models using cross-entropy loss, and then apply SCST and our method respectively. The same $RS*$ means they share the same pretrained model.

\begin{table*}[htbp]\small
  \centering
    \begin{tabular}{lcccccccc}
          & \multicolumn{1}{l}{Bleu1} & \multicolumn{1}{l}{Bleu2} & \multicolumn{1}{l}{Bleu3} & \multicolumn{1}{l}{Bleu4} & \multicolumn{1}{l}{ROUGE\_L} & \multicolumn{1}{l}{METEOR} & \multicolumn{1}{l}{CIDEr} & \multicolumn{1}{l}{SPICE} \\
    \midrule
    RS1+SCST & 80.3  & 64.5  & 49.9  & 38.0  & 58.3  & 28.4  & 127.2 & 21.9 \\
    RS1+Ours & \textbf{80.4} & \textbf{64.7} & \textbf{50.1} & \textbf{38.1} & \textbf{58.4} & \textbf{28.5} & \textbf{127.9} & \textbf{22.0} \\
    \midrule
    RS2+SCST & 80.2  & 64.5  & 49.9  & 37.9  & \textbf{58.3} & 28.3  & 127.2 & 21.9 \\
    RS2+Ours & \textbf{80.2} & \textbf{64.5} & \textbf{50.0} & \textbf{38.1} & 58.2  & \textbf{28.4} & \textbf{128.0} & \textbf{22.0} \\
    \midrule
    RS3+SCST & \textbf{80.2} & 64.5  & 50.0  & 38.1  & \textbf{58.3} & 28.3  & 127.3 & 21.8 \\
    RS3+Ours & 80.2  & \textbf{64.7} & \textbf{50.2} & \textbf{38.3} & 58.3  & \textbf{28.4} & \textbf{127.9} & \textbf{22.0} \\
    \midrule
    RS4+SCST & \textbf{80.2} & 64.5  & 49.9  & 37.9  & 58.2  & 28.3  & 127.0 & 21.8 \\
    RS4+Ours & 80.2  & \textbf{64.5} & \textbf{50.0} & \textbf{38.0} & \textbf{58.3} & \textbf{28.5} & \textbf{127.7} & \textbf{22.0} \\
    \midrule
    RS5+SCST & \textbf{80.2} & \textbf{64.6} & \textbf{50.2} & \textbf{38.4} & \textbf{58.4} & \textbf{28.5} & 127.6 & 21.9 \\
    RS5+Ours & 80.2  & 64.5  & 49.8  & 37.9  & 58.3  & 28.4  & \textbf{127.8} & \textbf{22.0} \\
    \midrule
    Mean(SCST) & 80.2  & 64.5  & 50.0  & 38.0  & 58.3  & 28.4  & 127.3 & 21.8 \\
    Mean(Ours) & \textbf{80.2} & \textbf{64.6} & \textbf{50.0} & \textbf{38.1} & \textbf{58.3} & \textbf{28.4} & \textbf{127.9} & \textbf{22.0} \\
    \bottomrule
    \end{tabular}%
  \caption{Within the first 5 block, the models share the same cross-entropy pretrained model (RS stands for random seed). The last block shows the average score of 5 models.}
  \label{tab:multipleruns}%
\end{table*}%

\paragraph{Training curves}

\begin{figure*}[t]
\centering
\includegraphics[width=0.24\linewidth]{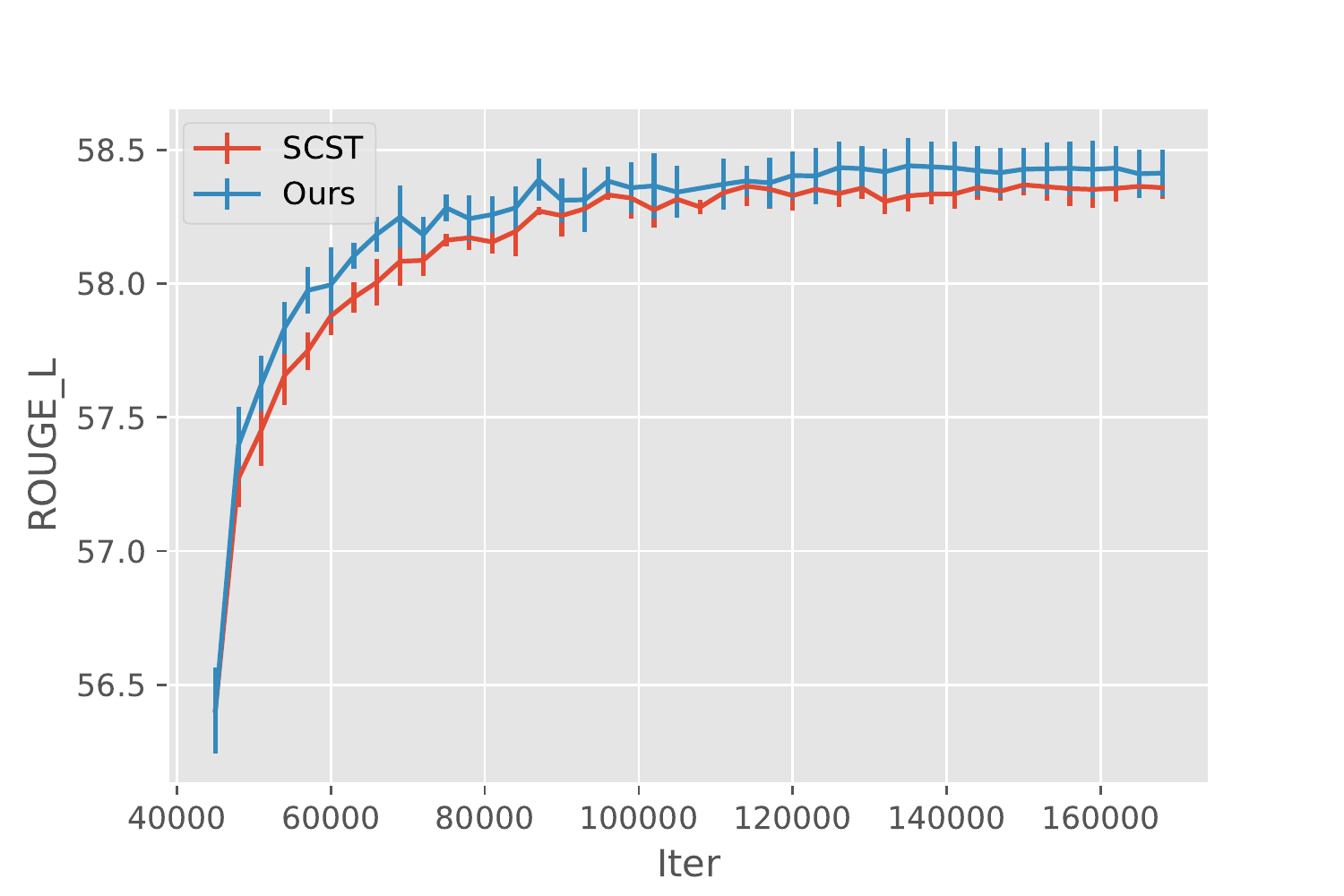}
\includegraphics[width=0.24\linewidth]{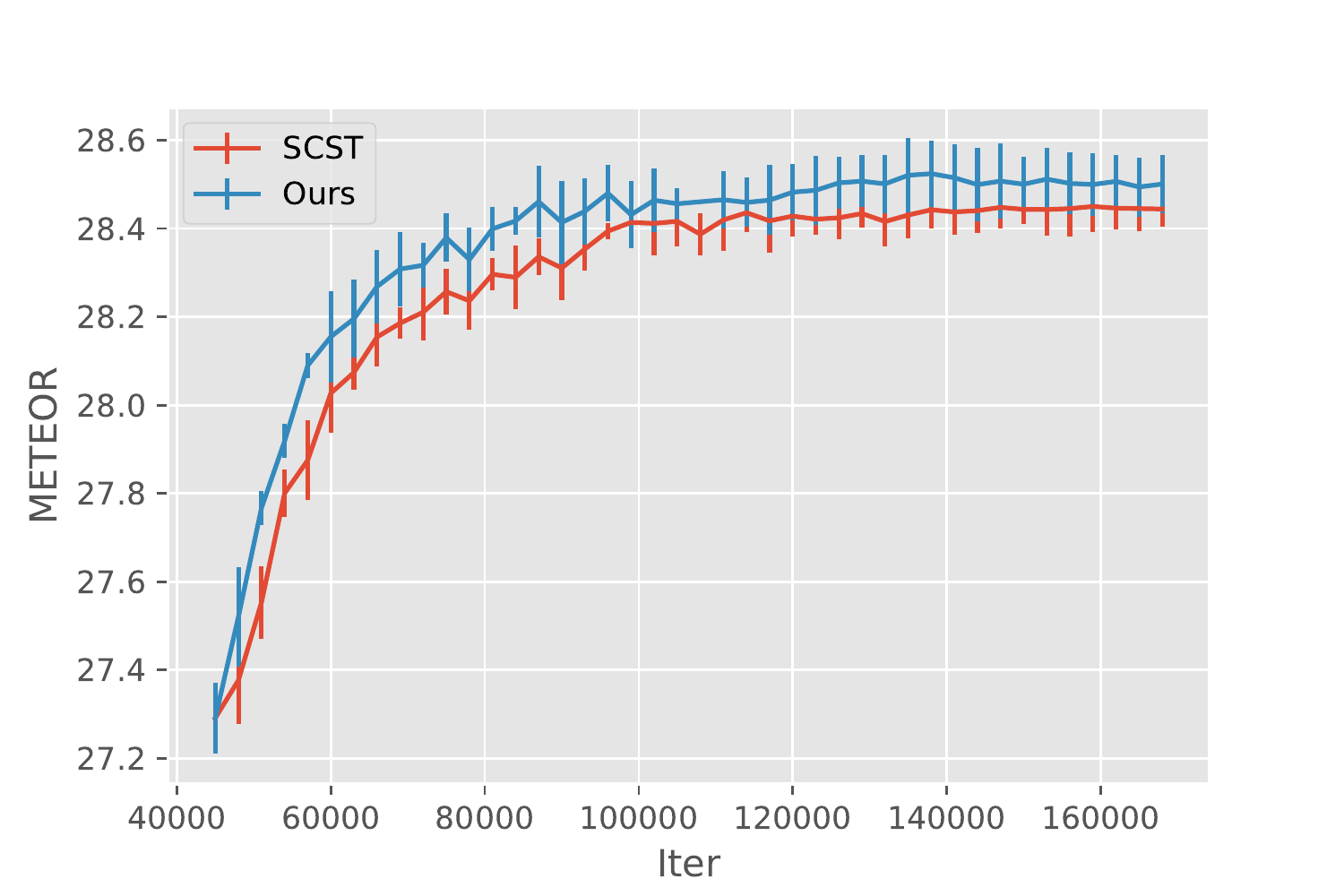}
\includegraphics[width=0.24\linewidth]{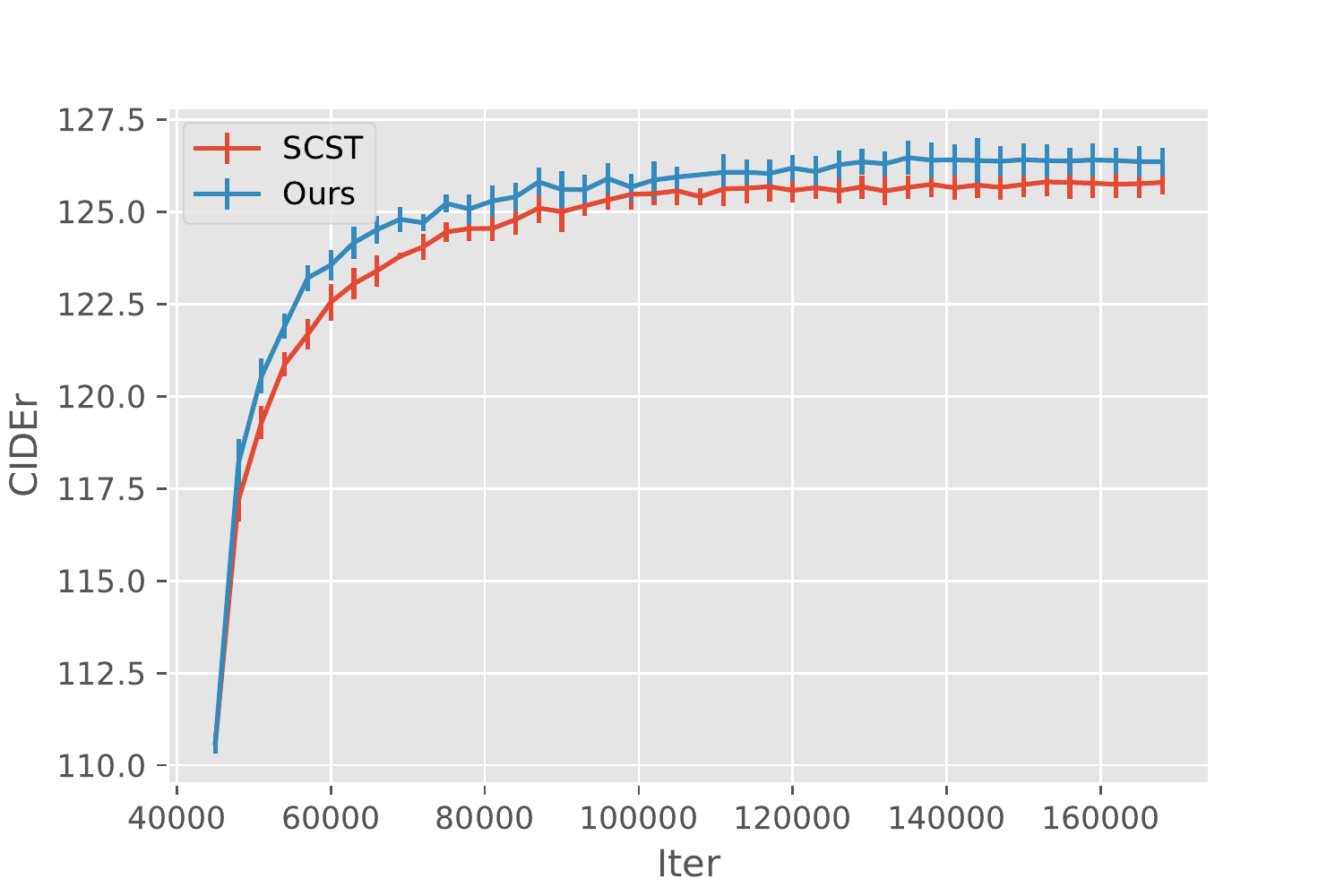}
\includegraphics[width=0.24\linewidth]{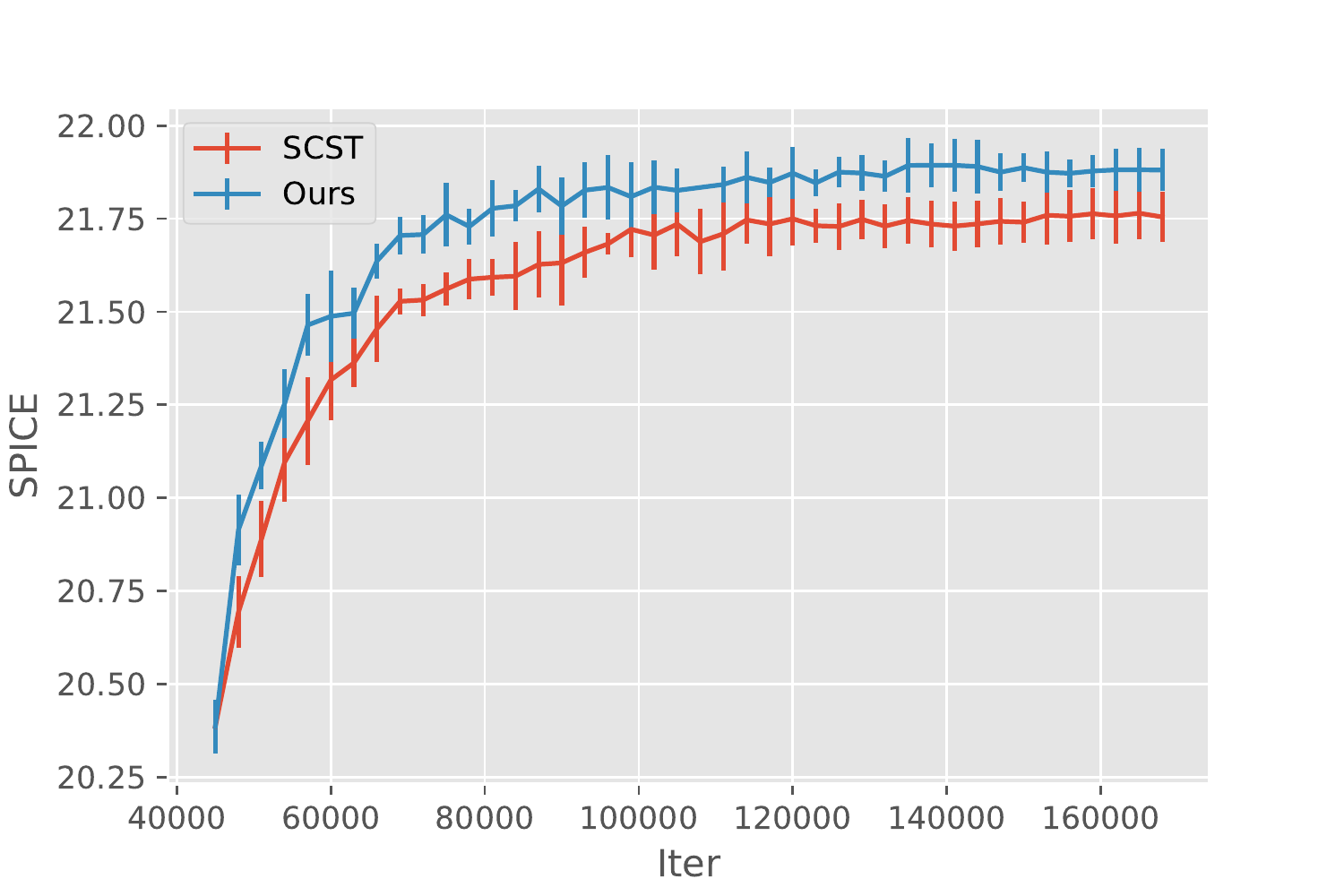}
\caption{Performance on validation set during training. (With UpDown(Long) + greedy decoding)}
\label{fig:training_curve}
\end{figure*}

Figure \ref{fig:training_curve} shows the model performance on the validation set during training, after entering Self-Critical stage. The scores are averaged over the 5 UpDown(Long) models above.

\paragraph{Is greedy decoding necessary for SCST}
We also experiment with a variant of SCST, by replacing the greedy decoding output with a sampled output. (This is similar to our method with $K=2$.)

Table \ref{tab:randombaseline} shows that one sample baseline is worse than greedy decoding. This is as expected, because using one sample to estimate the expected reward is too noisy, resulting in larger gradient variance, while the reward of greedy decoding output may be biased but more stable. It also shows that it is important to use sufficiently large $K$ to have a better estimation of expected reward.

\begin{table*}[htbp]\small
  \centering
    \begin{tabular}{lcccccccc}
          & \multicolumn{1}{c}{Bleu1} & \multicolumn{1}{c}{Bleu2} & \multicolumn{1}{c}{Bleu3} & \multicolumn{1}{c}{Bleu4} & \multicolumn{1}{c}{ROUGE\_L} & \multicolumn{1}{c}{METEOR} & \multicolumn{1}{c}{CIDEr} & \multicolumn{1}{c}{SPICE} \\
    \midrule
    UpDown(SCST) & 79.4  & 63.3  & 48.6  & \textbf{36.7} & 57.6  & \textbf{27.9} & \textbf{122.7} & \textbf{21.5} \\
    UpDown(Sample) & \textbf{79.6} & \textbf{63.5} & \textbf{48.7} & 36.7 & \textbf{57.7} & 27.8  & 122.1 & 21.3 \\
    \bottomrule
    \end{tabular}%
  \caption{Replacing the greedy decoding output $c^*$ in SCST with a separately drawn sample $\hat c^\prime$.}
  \label{tab:randombaseline}%
\end{table*}

% \subsection{Fix equivalent batch size}
% Here, we fix the number of samplesize * image size. 

% Table not yet.

% \subsection{Increase sample size}
% Here we fix the image batch size, and increase the sample size.

\paragraph{Variance reduction}
As stated in Sec. \ref{sec:nsc}, the motivation of using the average reward baseline is for better variance reduction. Here we show it indeed is better in practice.

The gradient variance is calculated as follows. At the end of each epoch, we take the saved model and run through the training set. We get the gradients from each training batch and calculate the variance for each parameter gradient across batches. To get a single value, we take the average of all the parameters. A mathematic expression of this process is:
$$V = \text{Mean}_i\left [\text{Var}_{b}[\text{grad}_{\theta_i}^b]\right ]$$

where $i$ is the index of each parameter; $b$ is the index of each batch; $\theta$ is the network parameters; $\text{grad}_{\theta_i}^b$ is the gradient of $\theta_i$ at batch $b$.

\begin{figure}
\centering
\includegraphics[width=0.75\linewidth]{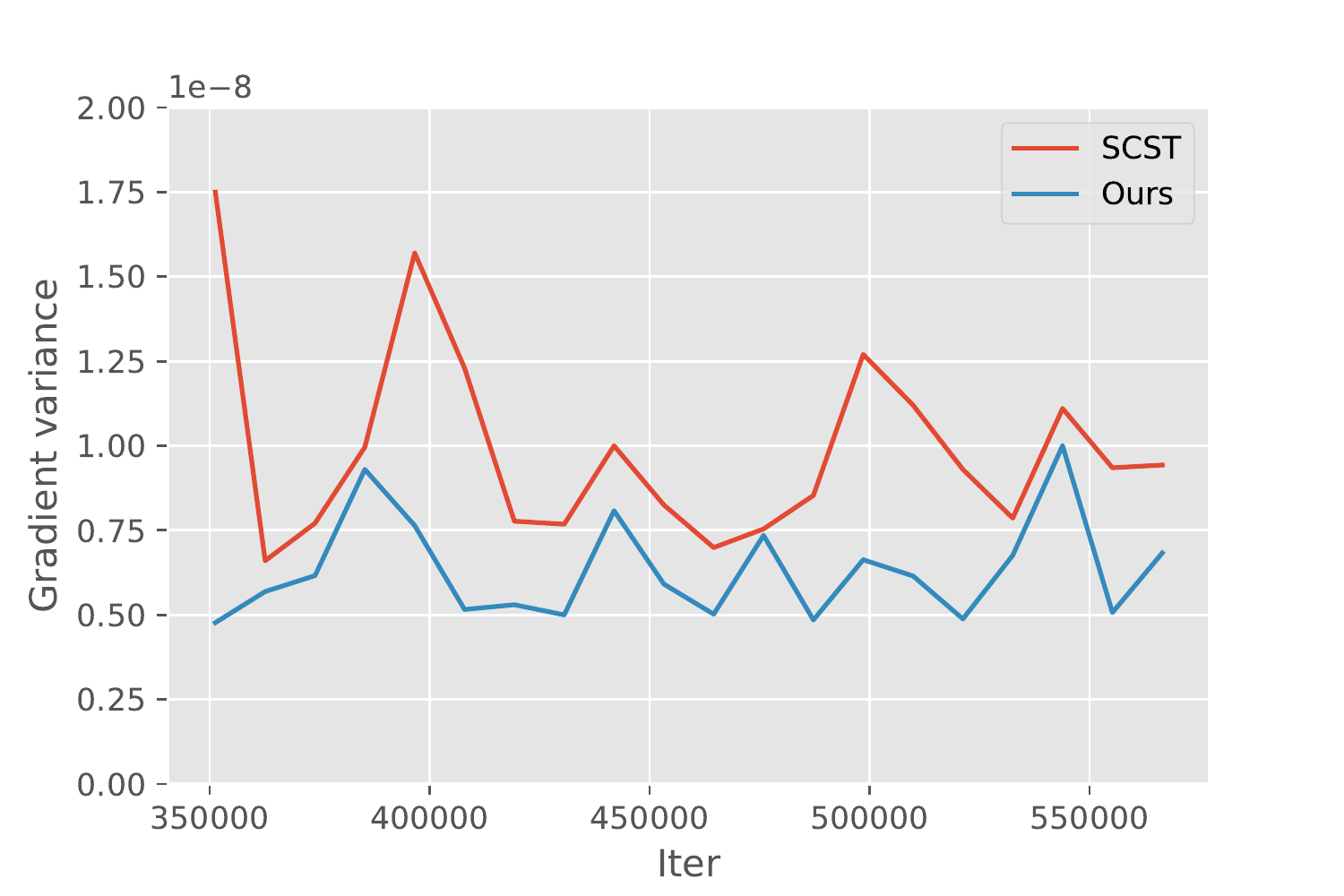}
\caption{The gradient variance on training set.(Model: UpDown)}
\label{fig:grad_var}
\end{figure}

As shown in Fig. \ref{fig:grad_var}, our method is always getting lower variance than SCST.

\section{Code release}
Code has been released at \url{https://github.com/ruotianluo/self-critical.pytorch}. More instructions of using this method are at \url{https://github.com/ruotianluo/self-critical.pytorch/tree/master/projects/NewSelfCritical}

\section{Conclusion}
We propose a variant of popular SCST, which can work as a drop-in replacement for SCST. This variant reduces the gradient variance when applying REINFORCE by modifying the baseline function. We show that this method is effective on Image Captioning task, and we believe it should benefit other tasks as well.

\bibliographystyle{acl_natbib}
\bibliography{ref}
\end{document}